\documentclass[11pt,a4paper]{article}
\usepackage[hyperref]{emnlp2020}
\usepackage{times}
\usepackage{latexsym}

\usepackage{microtype}
\usepackage{comment}
\usepackage{url}
\usepackage{amsmath}
\usepackage{amssymb}
\usepackage{algorithm}
\usepackage{graphicx} 
\usepackage{subfigure}
\usepackage[noend]{algpseudocode}
\usepackage{multirow}
\usepackage{wrapfig}
\usepackage{lipsum}
\usepackage{float}
\usepackage{mathtools}
\graphicspath{ {Figure/} }
\DeclareMathOperator*{\argmin}{\arg\!\min}

\aclfinalcopy 

\title{Enriching Word Embeddings with Temporal and Spatial Information}

\author{Hongyu Gong \quad Suma Bhat \quad Pramod Viswanath \\
  University of Illinois at Urbana-Champaign \\
  \texttt{\{hgong6,spbhat2,pramodv\}@illinois.edu}\\}
  
\date{}

\begin{document}
\maketitle
\begin{abstract}
 The meaning of a word is closely linked to sociocultural factors that can change over time and location, resulting in corresponding meaning changes. Taking a global view of words and their meanings in a  widely used language, such as English, may require us to capture more refined semantics for use in time-specific or location-aware situations, such as the study of cultural trends or language use. However, popular vector representations for words do not adequately include temporal or spatial information. In this work, we present a model for learning word representation  conditioned on time and location. In addition to  capturing meaning changes over time and location, we 
 require that the resulting word embeddings retain salient semantic  and geometric properties. 
 We train our model on time- and location-stamped corpora, and show using both quantitative and qualitative evaluations that it can  capture  semantics across time and locations. We note that our model compares favorably with the state-of-the-art for time-specific embedding, and serves as a new benchmark for location-specific  embeddings.
\end{abstract}

\section{Introduction}

The use of word embeddings as a form of lexical representation has transformed the use of natural language processing for many applications such as machine translation \cite{qi2018and} and language understanding \cite{peters2018deep}. 
The changing of word meaning over the course of time and space, termed \textit{semantic drift}, has been the subject of long standing research in diachronic linguistics \cite{ullmann1979semantics,blank1999new}.   {Additionally, the emergence of distinct geographically-qualified English varieties (e.g., South African English)    has  given rise to salient  lexical variation giving several English words different meanings depending on the geographic location of their use, as documented in studies on World Englishes \cite{kachru2006handbook,mesthrie2008world}}. Considering the multiplicity of meanings that a word can take over the span of time and space owing to inevitable linguistic, and sociocultural factors among others, a  static representation of a word as a single word embedding seems rather limited. Take the word \emph{apple} as an example. Its early to near-recent mentions in written documents referred only to a fruit, but in the recent times it is also the name of a large technology company. Another example is the title for the head of government, which is ``president'' in the USA, and is ``prime minister'' in Canada.

Naturally, we expect that one word should have different representations conditioned on the time or location. In this paper, we study how  word embeddings can be enriched to encode their semantic drift in time and space. Extending a recent line of research on time-specific embeddings, including the works by \citeauthor{bamler2017dynamic} and \citeauthor{yao2018dynamic},
we propose a model to capture varying lexical semantics across different conditions---of time and location. 

A key technical challenge of learning conditioned embeddings is to put the embeddings (derived from different time periods or geographical locations)  in the same vector space and preserve their geometry within and across different instances of the conditions.
Traditional approaches involve a two-step mechanism of first learning the sets of embeddings separately under the different conditions, and then aligning them via appropriate transformations  \cite{kulkarni2015statistically,hamilton2016diachronic,zhang2016past}.  {A primary limitation of these methods is their inadequate representation of word semantics, as we show in our comparative evaluation}. Another approach to conditioned embedding uses a loss function with regularizers over word embeddings across conditions for their smooth trajectory in the vector space \cite{yao2018dynamic}. {However, its scope is  limited to modeling semantic drift over only time.}

We propose a model for general conditioned embeddings, with the novelty  that it explicitly  preserves embedding geometry under different conditions and captures different degrees of word semantic changes. We summarize our contributions below. 
\begin{enumerate}
\item We propose an unsupervised model to learn condition-specific embeddings including time-specific and location-specific embeddings;
\item Using benchmark  datasets we demonstrate the state-of-the-art performance of the proposed model in accurately capturing word semantics across time periods and geographical regions;
\item We provide the first dataset\footnote{We release our data and code at \url{https://github.com/HongyuGong/EnrichedWordRepresentation}} to evaluate word embeddings across locations to foster research in this direction.
\end{enumerate}


\section{Related Works}
\label{sec:related}

\noindent\textbf{Time-specific embeddings}. The evolution of word meaning with time has been widely studied in sociolinguistics \cite{ullmann1979semantics,tang2018state}. 
Early approaches to uncovering these trends have relied on frequency based models which use frequency changes to trace semantic shift over time \cite{lijffijt2012ceecing,choi2012predicting,michel2011quantitative}. More recent works have sought to study these phenomena    using distributional models 
\cite{kutuzov2018diachronic,huang2019neural,schlechtweg2020semeval}. 

Existing approaches on time-specific embeddings can be divided into three categories: aligning independently trained embeddings across time, joint training of time-dependent embeddings and using contextualized vectors from pre-trained models. The first type of approaches include the work by \citeauthor{kulkarni2015statistically}, and that of \citeauthor{hamilton2016diachronic} and \citeauthor{zhang2016past}. They pre-trained multiple sets of embeddings for different times independently, and aligned one set of embedding with another set so that two sets of embeddings were comparable. 

The second approach---joint training---aims to guarantee the alignment of embeddings in the same vectors space so that they are directly comparable. Compared with the previous category of approaches, the
joint learning of time-stamped embeddings has shown improved abilities to capture semantic changes across time. 
\citeauthor{bamler2017dynamic} used a probabilistic model to learn time-specific embeddings \cite{bamler2017dynamic}. They put Gaussian assumption on the evolution of embeddings to guarantee the embedding alignment. \citeauthor{yao2018dynamic} learned embeddings by the factorization of positive pointwise mutual information (PPMI) matrix. They imposed L2 constraints on embeddings from neighboring time periods for embedding alignment \cite{yao2018dynamic}. \citeauthor{rosenfeld2018deep} proposed a neural model to first encode time and word information respectively and then to learn time-specific embeddings \cite{rosenfeld2018deep}. \citeauthor{dubossarsky2019time} aligned word embeddings by sharing their context embeddings at different times \cite{dubossarsky2019time}.

Some recent works fall in the third category, retrieving contextualized representations from pre-trained models such as BERT \cite{devlin2018bert} as time-specific sense embeddings of words \cite{hu2019diachronic,DBLP:conf/acl/GiulianelliTF20}. These pre-trained embeddings are limited to the scope of local contexts, while we learn the global representation of words in a given time or location. 

The underlying mathematical models of these previous works on temporal embeddings are discussed in the supplementary material. Our model belongs to the second category of joint embedding training. Different from previous works, our embedding is based on a model which explicitly takes into account the important semantic properties of time-specific embeddings.

\noindent\textbf{Embedding with spatial information}.
Lexical semantics is also sensitive to spatial factors. For example, the word denoting the regional head may be used differently depending on the region. For instance, the word may refer to \emph{president}, or \textit{prime minister} or \textit{king} depending on the region. 
Language variation across regional contexts has been analyzed in sociolinguistics and dialectology studies (e.g.,\cite{silva2006analyzing,kulkarni2016freshman}). It is also understood that a deeper  understanding of semantics enhanced with location information is critical to location-sensitive applications such as  content localization of global search engines \cite{brandon2001localization}. 

Some approaches towards this include, a latent variable model  proposed for geographical linguistic variation \cite{eisenstein2010latent} and a skip-gram model for geographically situated language \cite{bamman2014distributed}.
The current study is most similar to  \cite{bamman2014distributed} with the overlap in our intents to learn location-specific embeddings for measuring semantic drift. 
Most studies on location-dependent language resort to a qualitative evaluation, whereas  \cite{bamman2014distributed} resorts to a  quantitative analysis for entity similarity. However, it is limited to a given region without exploring semantic equivalence of words across different geographic regions. To the extent we are aware, this is the first study to present a  quantitative evaluation of word representations across geographical regions with the use of  a dataset. 

\section{Model}
\label{sec:model}
We now introduce the model on which the condition-specific embedding training is based in this section. We assume access to a corpus divided into sub-corpora based on their conditions (\textit{time} or \textit{location}), and texts in the same condition (e.g., same time period) are gathered in each sub-corpus. For each condition, the co-occurrence counts of word pairs gathered from its sub-corpus are the corpus statistics we use for the embedding training. We note that because these sub-corpora vary in size,  we scale the  word co-occurrences of every condition so that all sub-corpora have the same total number of word pairs. We term the scaled value of word co-occurrences of word $w_{i}$ and $w_{j}$ in condition $c$ as $\mathbf{X}_{i,j,c}$.

A static model (without regard to the temporal or spatial conditions) proposed by \citeauthor{arora2015rand} provides the unifying theme for the seemingly different embedding approaches of word2vec and GloVe. In particular, It reveals that corpus statistics such as word co-occurrences could be estimated from embeddings. Inspired by this, we proposed a model for conditioned embeddings, and characterize such a model by its ability to capture the lexical semantic properties across different conditions.

\subsection{Properties of Conditioned Embeddings}
\label{subsec:property}
Before exploring the details of our  model for condition-specific embeddings, we  discuss some desired semantic properties of these embeddings. We expect the embeddings to capture time- and location-sensitive lexical semantics. We denote by $c$ the condition we use to refine word embeddings, which can be a specific time period or a location. We then have temporal embeddings if the condition is \textit{time period}, and spatial embeddings if the condition is \textit{location}. For a word $w$, the condition-specific word embedding for condition $c$ is denoted as $\mathbf{v}_{w,c}$. The key semantic properties of the 
condition-specific word embedding, which we consider in our model are: \\
\noindent(1) \textbf{Preservation of geometry}. One geometric property of static embeddings is that the difference vector encodes word relations, i.e., $\mathbf{v}_{\text{bigger}}-\mathbf{v}_{\text{big}} \approx \mathbf{v}_{\text{greater}}-\mathbf{v}_{\text{great}}$ \cite{mikolov2013distributed}. Analogously,  
for the condition-specific embedding of semantically stable words across conditions, given word pairs $(w_{1},w_{2})$ and $(w_{3},w_{4})$ with the same underlying lexical relation, we expect the following equation to hold in any condition $c$.
   \begin{align}
   \mathbf{v}_{w_{1},c}-\mathbf{v}_{w_{2},c} \approx \mathbf{v}_{w_{3},c} - \mathbf{v}_{w_{4},c}.
   \end{align}
This property is implicitly preserved in approaches aligning independently trained embeddings with linear transformations \cite{kulkarni2015statistically}.\\
\noindent(2) \textbf{Consistency over conditions}. Most word meanings change slowly over a given condition, i.e., their condition-specific word embeddings should be highly correlated \cite{hamilton2016diachronic}. 
When the condition is time period, for example, $c_{1}$ is the year 2000, and $c_{2}$ is the year 2001,  we expect that for a given word, $\mathbf{v}_{w,c_{1}}$ and $\mathbf{v}_{w,c_{2}}$ have high similarity given their temporal proximity. {The consistency property is preserved in models which jointly train embeddings across conditions (e.g., \cite{yao2018dynamic}).} \\
\noindent(3) \textbf{Different degrees of word change}. Although word meanings change over time, not all words undergo this change to the same degree; some words change dramatically while others stay relatively stable across conditions \cite{blank1999new}. In our formulation, we require the representation to capture the different degrees of word meaning change. {This property is unexplored in prior studies.} 

{We incorporate these semantic properties as explicit constraints into our model for condition-specific embeddings, which we formulate as an optimization problem. }

\subsection{Model}
We propose a model that generates embeddings satisfying the semantic properties as discussed above. Writing the embedding $\mathbf{v}_{w,c}$ of word $w$ in condition $c$ as a function of its condition-independent representation $\mathbf{v}_{w}$, condition representation vector $\mathbf{q}_{c}$ and deviation embedding $\mathbf{d}_{w,c}$:
\begin{align}
\label{eq:vector}
\mathbf{v}_{w,c} = \mathbf{v}_{w}\odot \mathbf{q}_{c}+ \mathbf{d}_{w,c},
\end{align}
where $\odot$ is Hadamard product (i.e., elementwise multiplication). We decompose the conditioned representation into three component embeddings. {This novel representation is motivated by the intuition that a word $w$ usually carries its basic meaning $\mathbf{v}_{w}$ and its meaning is influenced by different conditions represented by $\mathbf{q}_{c}$. Moreover, words have different degrees of meaning variation, which is captured by the deviation embedding $\mathbf{d}_{w,c}$.} 

We begin with a model proposed by \citeauthor{arora2015rand} for static word embeddings regardless of the temporal or spatial conditions \cite{arora2015rand}.
Let $\mathbf{v}_{w}$ be the static representation of word $w$.  For a pair of words $w_{1}$ and $w_{2}$, the static model assumes that
\begin{align}
\label{eq:static_vector}
\log\mathbb{P}(w_{1},w_{2}) \approx \frac{1}{2}\lVert \mathbf{v}_{w_{1}} + \mathbf{v}_{w_{2}}\rVert^{2},
\end{align}
where $\mathbb{P}(w_{1},w_{2})$ is the co-occurrence probability of these two words in the training corpus.

Let $\mathbb{P}_{c}(w_{1},w_{2})$ be the co-occurrence probability of word pair $(w_{1}, w_{2})$ in the condition $c$. Based on the static model in Eq.~(\ref{eq:static_vector}), for a condition $c$ we have
\begin{align}
\label{eq:gen}
\log\mathbb{P}_{c}(w_{1},w_{2}) \approx \frac{1}{2}\lVert \mathbf{v}_{w_{1},c} + \mathbf{u}_{w_{2},c}\rVert^{2}.
\end{align}

{Here, borrowing ideas from previous embedding algorithms including word2vec \cite{mikolov2013distributed} and GloVe \cite{pennington2014glove},  we use two sets of word embeddings $\{\mathbf{v}_{w,c}\}$ and $\{\mathbf{u}_{w,c}\}$ for a word $w_{1}$ and its context word $w_{2}$ respectively in condition $c$. 
Accordingly, we have two sets of condition-independent embeddings $\{\mathbf{v}_{w}\}$ and $\{\mathbf{u}_{w}\}$, and two sets of deviation vectors $\{\mathbf{d}_{w_,c}\}$ and $\{\mathbf{d}'_{w,c}\}$.}
The condition-specific embeddings in Eq.~(\ref{eq:vector}) can be written as:
\begin{align}
\label{eq:two_vectors}
\left\{
\begin{array}{cc}
    \mathbf{v}_{w_{1},c} &= \mathbf{v}_{w_{1}}\odot\mathbf{q}_{c} + \mathbf{d}_{w_{1},c}  \\
    \mathbf{u}_{w_{2},c} &= \mathbf{u}_{w_{2}}\odot\mathbf{q}_{c} + \mathbf{d}'_{w_{2},c} 
\end{array}
\right.
\end{align}

By combining Eq.~(\ref{eq:gen}) and (\ref{eq:two_vectors}), we derive the model for condition-specific embeddings:
\begin{align}
\nonumber
\log\mathbb{P}_{c}(w_{1},w_{2}) &\approx \frac{1}{2}\lVert\mathbf(\mathbf{v}_{w_{1}}\odot\mathbf{q}_{c}
+\mathbf{d}_{w_{1},c}) \\
&+ (\mathbf{u}_{w_{2}}\odot \mathbf{q}_{c}+\mathbf{d}_{w_{2},c}')\rVert^{2}.
\end{align}

\noindent This model can be simplified as
\begin{align}
\nonumber
&\log\mathbb{P}_{c}(w_{1},w_{2}) \approx b_{w_{1},c} + b_{w_{2},c}' + \\
\label{eq:logprob}
&(\mathbf{v}_{w_{1}}\odot\mathbf{q}_{c}+\mathbf{d}_{w_{1},c})^{T}(\mathbf{u}_{w_{2}}\odot \mathbf{q}_{c}+\mathbf{d}_{w_{2},c}'),
\end{align}


where $b_{w_{1},c}$ and $b_{w_{2},c}'$ are bias terms introduced to replace the terms $\lVert \mathbf{v}_{w_{1},c}\rVert^{2}$ and $\lVert \mathbf{u}_{w_{2},c}\rVert^{2}$ respectively. 
We document the derivation details of Eq.~(\ref{eq:logprob})  in the supplementary material.

\noindent\textbf{Optimization problem}.  This model enables us to use the conditioned embeddings to estimate the word co-occurrence probabilities in a specific condition. Conversely, we can formulate an optimization problem to train the conditioned embeddings from the word co-occurrences based on our model.

We count the co-occurrences of all  word pairs ($w_{1}$, $w_{2}$) in different conditions based on the respective sub-corpora. For example, we count word co-occurrences over different time periods to incorporate temporal information into word embeddings, and we count word pairs in different locations to learn spatially sensitive word representations.

Recall that $\mathbf{X}_{i,j,c}$ is the scaled co-occurrence counts  of $w_{i}$ and $w_{j}$ in condition $c$. 
Denote by $W$ the total vocabulary and by $C$ the number of conditions, where $C$ is the number of time bins for the temporal condition 
or the number of locations for the location condition. 
 Suppose that $\mathbf{V}$ is an $(m\times|W|)$ condition-independent word embedding matrix, where each column corresponds to an $m$-dimension word vector $\mathbf{v}_{w}$. Matrix $\mathbf{U}$ is an $(m\times|W|)$ basic context embedding matrix with each column as a context word vector $\mathbf{u}_{w}$. Matrix $\mathbf{Q}$ is an $(m\times C)$ matrix, where each column is a condition vector $\mathbf{q}_{c}$. As for deviation matrices, $\mathbf{D}_{m\times |W|\times C}$ and $\mathbf{D}'_{m\times|W|\times C}$ consist of $m$-dimension deviation vectors $\mathbf{d}_{w,c}$ and $\mathbf{d}_{w,c}'$ respectively for word $w$ in condition $c$.

{Our goal is to learn embeddings $\mathbf{U}$, $\mathbf{Q}$ and $\mathbf{D}$ so as to approximate the word co-occurrence counts based on the model in Eq.(\ref{eq:logprob}).
Here, we design a loss function 
to be the approximation error of the embeddings, which is  the mean square error between the condition-specific co-occurrences counted from the respective sub-corpora and their estimates from the embeddings.}

To satisfy the property 2 of condition-specific embeddings, we impose $L_{2}$ constraints $\lVert \mathbf{q}_{a}-\mathbf{q}_{b}\rVert^{2}$ on the embeddings of condition $a$ and $b$ to guarantee the consistency over conditions. For time-specific embeddings, the constraints are for adjacent time bins. 
As for location-sensitive embeddings, the constraints are for all pairs of location embeddings. 

Furthermore, to account for the slow change in meaning of most words across conditions (as in time periods or locations) listed as property 3 of conditioned embeddings,  we also include $L_{2}$ constraints $\lVert\mathbf{D}\rVert^{2}$ and $\lVert\mathbf{D}'\rVert^{2}$ on the deviation terms to penalize big changes. 

Putting together the approximation error, constraints on condition embeddings and deviations, we have the following loss function: 

\begingroup
\small
\begin{align}
\nonumber
L = &\sum\limits_{c=1}^{C}\sum\limits_{i=1}^{|W|}\sum\limits_{j=1}^{|W|} \left((\mathbf{V}_{i}\odot \mathbf{Q}_{c} + \mathbf{D}_{i,c})^{T}(\mathbf{U}_{j}\odot \mathbf{Q}_{c}+\mathbf{D}_{j,c}')\right.\\
\nonumber
&\left.+\mathbf{b}_{i,c}+\mathbf{b}_{j,c}'-\log(\mathbf{X}_{i,j,c})\right)^{2} \\
\label{eq:mse}
&+ \frac{\alpha}{2}\sum\limits_{a,b}\lVert \mathbf{Q}_{a}-\mathbf{Q}_{b}\rVert^{2} + \frac{\beta}{2}(\lVert\mathbf{D}\rVert^{2} + \lVert\mathbf{D}'\rVert^{2}).
\end{align}
\endgroup
In addition to ensuring a smooth  trajectory of the embeddings, the penalization on the deviations $\mathbf{D}$ and $\mathbf{D}'$ is necessary to avoid the degenerate case that $\mathbf{Q}_{c}=\bf{0}, \forall c$.



We note that, for the constraint on condition embeddings in the loss function $L$, for time-specific embeddings we use  $\sum\limits_{c=1}^{C-1}\lVert \mathbf{Q}_{c+1}-\mathbf{Q}_{c}\rVert^{2}$, whereas for location-specific embeddings, the constraint becomes $\sum\limits_{a=1}^{C-1}\sum\limits_{b=a+1}^{C}\lVert \mathbf{Q}_{a}-\mathbf{Q}_{b}\rVert^{2}$.


\noindent\textbf{Model Properties}. We have presented our approach to learning conditioned embeddings. Now we will show that the proposed model satisfies the aforementioned key properties in Section \ref{subsec:property}. We start with the property of geometry preservation. For a set of semantically stable words $S=\{w_{1}, w_{2}, w_{3},w_{4}\}$, it is known that $d_{w,c}\approx0$ for $w\in S$. Suppose that the relation between $w_{1}$ and $w_{2}$ is the same as the relation between $w_{3}$ and $w_{4}$, i.e., $\mathbf{v}_{w_{1}}-\mathbf{v}_{w_{2}} = \mathbf{v}_{w_{3}} - \mathbf{v}_{w_{4}}.$ Given Eq.~(\ref{eq:vector}) for any condition $c$, it holds that
\begin{align}
\nonumber
&\mathbf{v}_{w_{1},c}-\mathbf{v}_{w_{2},c} \approx (\mathbf{v}_{w_{1}}-\mathbf{v}_{w_{2}})\odot q_{c} \\
&\approx(\mathbf{v}_{w_{3}}-\mathbf{v}_{w_{4}})\odot q_{c} \approx \mathbf{v}_{w_{3},c} - \mathbf{v}_{w_{4},c}.
\end{align}

As for the second property of consistency over conditions, we again consider a stable word $w$. Its conditioned embedding $v_{w,c}$ in condition $c$ can be written as $\mathbf{v}_{w,c}=\mathbf{v}_{w}\odot \mathbf{q}_{c}$. As is shown in Eq.~(\ref{eq:mse}), the $L_2$ constraint $\lVert \mathbf{q}_{a}-\mathbf{q}_{b}\rVert^{2}$ is put on different condition embeddings. The difference between word embeddings of $w$ under two conditions $a$ and $b$ are:
\begin{align}
\nonumber
\lVert \mathbf{v}_{w,a} - \mathbf{v}_{w,b}\rVert^{2} &= \lVert \mathbf{v}_{w}\odot(\mathbf{q}_{a}-\mathbf{q}_{b})\rVert^{2} \\
&\leq \frac{1}{2}\lVert \mathbf{v}_{w}\rVert^{2}\cdot \lVert \mathbf{q}_{a}-\mathbf{q}_{b}\rVert^{2}.
\end{align}
According to Cauchy-Schwartz inequality, the $L_2$ constraint on condition vectors $\mathbf{q}_{a}-\mathbf{q}_{b}$ also acts as a constraint on word embeddings. With a large coefficient $\alpha$, it prevents the embedding from differing too much across conditions, and guarantees the smooth trajectory of words.

Lastly we show that our model captures the degree of word changes. The deviation vector $\mathbf{d}_{w,c}$ we introduce in the model captures such changes.
{The $L_2$ constraint on $\lVert \mathbf{d}_{w,c}\rVert$ shown in Eq.~(\ref{eq:mse}) forces small deviation on most words which are smoothly changing across conditions. We assign a small coefficient $\beta$ to this constraint to allow sudden meaning changes in some words. The hyperparameter setting is discussed below.}

\noindent\textbf{Embedding training}. We have hyperparameters $\alpha$ and $\beta$ as weights on the word consistency and the deviation constraints. We set $\alpha=1.5$ and $\beta=0.2$ in time-specific embeddings, and $\alpha=1.0$ and $\beta=0.2$ in location-specific embeddings.

At each training step, we randomly select a nonzero element $x_{i,j,c}$ from the co-occurrence tensor $\mathbf{X}$. Stochastic gradient descent with adaptive learning rate is applied to update $\mathbf{V}$, $\mathbf{U}$, $\mathbf{Q}$, $\mathbf{D}$, $\mathbf{D}'$, $\mathbf{d}$ and $\mathbf{d}'$, which are relevant to $x_{i,j,c}$ to minimize the loss $L$. 
The complexity of each step is $O(m)$, where $m$ is the embedding dimension. In each epoch, we traverse all nonzero elements of $\mathbf{X}$. Thus we have $\text{nnz}(\mathbf{X})$ steps where $\text{nnz}(\cdot)$ is the number of nonzero elements. Although $\mathbf{X}$ contains $O(|W|^{2})$ elements, $\mathbf{X}$ is very sparse since many words do not co-occur, so $\text{nnz}(\mathbf{X})\ll |W|^{2}$. The time complexity of our model is $O(E\cdot m\cdot\text{nnz}(\mathbf{X}))$ for $E-$epoch training. We set $E=40$ in training both temporal and spatial word embeddings.

\noindent\textbf{Postprocessing}. 
We note that embeddings under the same condition are not centered, i.e., the word vectors are distributed around some non-zero point. We center these vectors by removing the mean vector of all embeddings in the same condition. 
The centered embedding $\tilde{\mathbf{v}}_{w,c}$ of word $w$ under condition $c$ is:
\begin{align}
\tilde{\mathbf{v}}_{w,c} = \mathbf{v}_{w,c} - \frac{1}{|W|}\sum\limits_{\bar{w}\in W} \mathbf{v}_{\bar{w},c}.
\end{align}

The similarity between words across conditions is measured by the cosine similarity of their centered embeddings $\{\tilde{\mathbf{v}}_{w,c}\}$.

\begin{table*}[htbp!]
\centering
\begin{tabular}{|c|l|}
\hline
Across time & \begin{tabular}[c]{@{}l@{}}the, in, to, a, of, it, by, with, at, was, are, and, on, who, for, not, they, but, he, is, from, \\have, as, has, their, about, her, been, there, or, will, this, said, would 
\end{tabular}  \\ \hline
Across regions & \begin{tabular}[c]{@{}l@{}}in, from, at, could, its, which, out, but, on, all, has, so, is, are, had, he, been, by, an, it, \\as, for, was, this, his, be, they, we, her, that, and, with, a, of, the
\end{tabular} \\ \hline
\end{tabular}
\caption{Stable Words across Time and Locations}
\label{table:stable_words}
\end{table*}

\section{Experiments}
\label{sec:exp}
In this section, we compare our condition-specific word embedding models with corresponding  state-of-the-art models combined with temporal or spatial information. The dimension of all vectors is set as $50$. We have the following baselines:\\
\noindent(1) \textbf{Basic word2vec} (BW2V). It is word2vec CBOW model, which is trained on the entire corpus without considering any temporal or spatial partition \cite{mikolov2013distributed}; \\
\noindent(2) \textbf{Transformed word2vec} (TW2V). Multiple sets of embeddings are trained separately for each condition. Two sets of embeddings are then aligned via a linear transformation
\cite{kulkarni2015statistically}.\\
\noindent(3) \textbf{Aligned word2vec} (AW2V): Similar to TW2V, sets of embeddings are first trained independently and then aligned via orthonormal transformations \cite{hamilton2016diachronic}.\\
\noindent(4) \textbf{Dynamic word embedding} (DW2V): This approach proposes a joint training of word embeddings at different times with alignment constraints on temporally adjacent sets of embeddings \cite{yao2018dynamic}. We modify this baseline for location based embeddings by putting its alignment constraints on every two sets of embeddings.


\subsection{Training Data}
We used two corpora as training data--the time-stamped news corpus of the New York Times collected by \cite{yao2018dynamic} to train time-specific embeddings and  a collection of location-specific texts in English, provided by the International Corpus of English project \cite{ICE} for location-specific embeddings.

\textbf{New York Times corpus}. The news dataset from New York Times consists of $99,872$ articles from 1990 to 2016. We use time bins of size one-year, and divide the corpus into 27 time bins.

\textbf{International Corpus of English (ICE)}. The ICE project collected written and spoken material in English (one million words each) from different regions of the world after 1989. We used the written portions collected from  Canada, East Africa, Hong Kong, India, Ireland, Jamaica, the Philippines, Singapore and the United States of America. 

Deviating from previous works, which remove both stop words and infrequent words from the vocabulary \cite{yao2018dynamic}, we only remove words with observed frequency count less than a threshold. {We keep the stop words to show that the trained embedding is able to identify them as being semantically stable.}
The frequency threshold is set to $200$ (the same as \cite{yao2018dynamic}) for the New York Times corpus, and to $5$ for the ICE corpus given that the smaller size of ICE corpus results in lower word frequency than the news corpus.

We evaluate the enriched word embeddings for the following aspects:
\begin{enumerate}
\item \textbf{Degree of semantic change}. As mentioned in the list of desired properties of conditioned embeddings, words undergo semantic change to different degrees. We check whether our embeddings can identify words whose meanings are relatively stable across conditions. These stable words will be discussed as part of  the qualitative evaluation.
\item \textbf{Discovery of semantic change}. Besides stable words, we also study words whose meaning changes drastically over conditions. Since a word's neighbors in the embedding space can reflect its meaning, we find the neighbors in different conditions to demonstrate how the word meaning changes. The discovery of semantic changes will be discussed as part of our qualitative evaluation.
\item \textbf{Semantic equivalence across conditions}. All condition-specific embeddings are expected to be in the same vector space, i.e., the cosine similarity between a pair of embeddings  reflects their lexical similarity even though they are from different condition values. {Finding semantic equivalents with the derived embeddings will be discussed in the quantitative evaluation.} 
\end{enumerate}

\begin{figure*}[h]
\centering
\begin{minipage}{0.51\textwidth}
\centerline{\includegraphics[width=\linewidth]{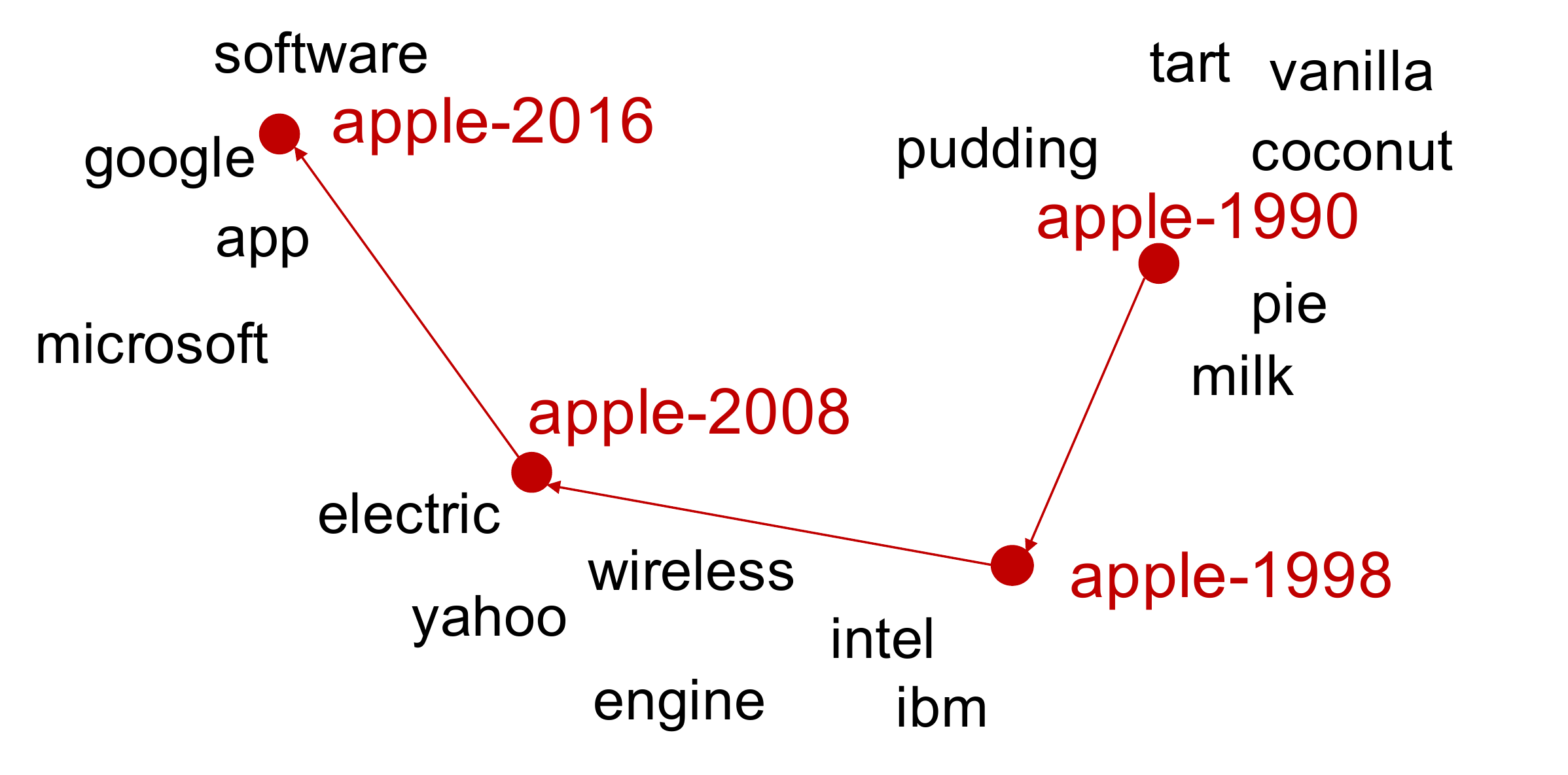}}
\centerline{\small{(a) Word ``apple'' and its neighbors across time.}}
\label{fig:apple_trajectory}
\end{minipage}
\begin{minipage}[c]{0.48\textwidth}
\centerline{\includegraphics[width=\linewidth]{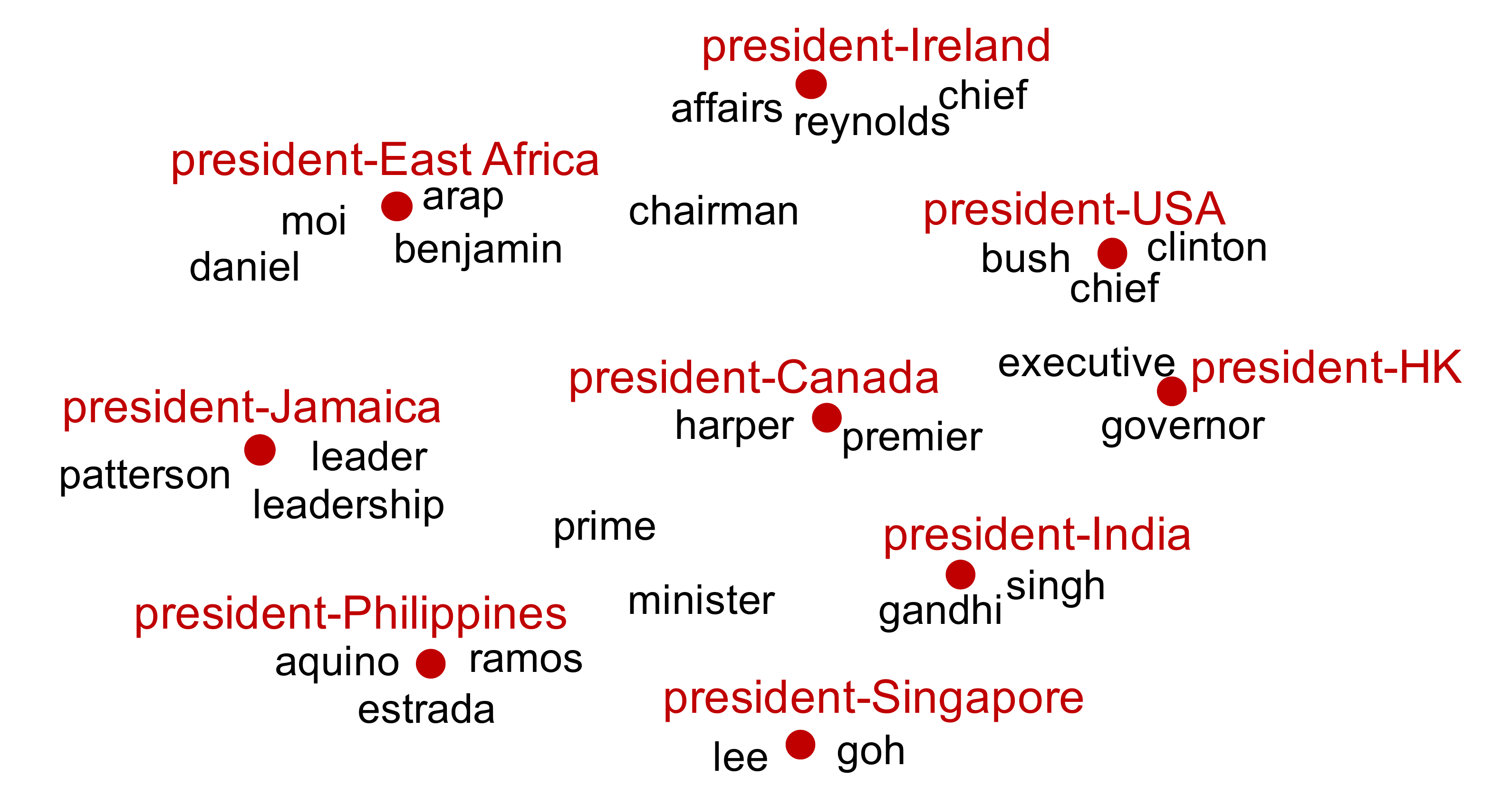}}
\centerline{\small{(b) Word ``president'' and its neighbors across locations.}}
\label{fig:president_trajectory}
\end{minipage}
\caption{The trajectory of word embeddings over time and locations.}
\label{fig:trajectory}
\end{figure*}

\subsection{Qualitative Evaluation}

We first identify words that are semantically stable across time and locations respectively. {Cosine similarity of embeddings reflects the semantic similarity of words. The embeddings of stable words should have high similarity across conditions since their semantics do not change much with conditions.}
Therefore, we average the cosine similarity of words between different time durations or locations as the measure of word stability, and rank the words in terms of their stability. The most stable words are listed in Table~\ref{table:stable_words}.
We notice that a vast majority of these stable words are frequent words such as function words. It may be interpreted based on the fact that these are words that encode structure \cite{gong2017prepositions,gong2018embedding}, and that the structure of well-edited English text has not changed much across time or locations \cite{poirier2014method}.
It is also in line  with our general linguistic knowledge; function words are those with high frequency in corpora, and are semantically relatively stable  \cite{hamilton2016diachronic}.

Next we focus on the words whose meaning varies with time or location.
We first evaluate the semantic changes of embeddings trained on time-stamped news corpus, and choose the word \emph{apple} as an example (more examples are included in the supplementary material). We plot the trajectory of the embeddings of {\emph{apple} and its semantic neighbors} over time in Fig.~\ref{fig:trajectory}(a). These word vectors are projected to a two-dimensional space using the locally linear embedding approach \cite{roweis2000nonlinear}. 
We notice that the word \emph{apple} usually referred to a fruit in 1990 given that its neighbors are food items such as \emph{pie} and \emph{pudding}. In recent years, the word has taken on the sense of the technology company Apple, which can be seen from the fact that \emph{apple} is close to words denoting technology companies such as \emph{google} and \emph{microsoft} after 1998.

{We also evaluate the location-specific word embeddings trained on the ICE corpus on the task of semantic change discovery. Take the word \emph{president} as an example. We list its neighbors in different locations in Fig.~\ref{fig:trajectory}(b). It is close to names of the regional leaders. The neighbors are president names such as \emph{bush} and \emph{clinton} in USA, and prime minister names such as \emph{harper} in Canada and \emph{gandhi} in India. }
This suggests that the embeddings are qualitatively shown to capture semantic changes across different conditions. 

\begin{table*}[htbp!]
\centering
\resizebox{1.0\textwidth}{!}{
\begin{tabular}{|c|c|c|c|c|c|c|c|c|c|c|}
\hline
Dataset & \multicolumn{5}{c|}{Temporal testset 1} & \multicolumn{5}{c|}{Temporal testset 2} \\ \hline
Metric & MRR & MP@1 & MP@3 & MP@5 & MP@10 & MRR & MP@1 & MP@3 & MP@5 & MP@10 \\ \hline
BW2V & 0.36 & 0.27 & 0.42 & 0.48 & 0.56 & 0.05 & 0.00 & 0.08 & 0.08 & 0.20  \\ \hline
TW2V & 0.09 & 0.05 & 0.12 & 0.15 & 0.19 & 0.07 & 0.04 & 0.08 & 0.10 & 0.14 \\ \hline
AW2V & 0.16 & 0.11 & 0.18 & 0.22 & 0.30 & 0.05 & 0.02 & 0.05 & 0.08 & 0.14  \\ \hline
DW2V & 0.42 & 0.33 & 0.49 & \textbf{0.55} & \textbf{0.62} & \textbf{0.14} & \textbf{0.08} & \textbf{0.16} & \textbf{0.22} & \textbf{0.38} \\ \hline
CW2V & \textbf{0.43} & \textbf{0.34} & \textbf{0.51} & \textbf{0.55} & \textbf{0.62} & 0.13 & \textbf{0.08} & \textbf{0.16} & 0.19 & 0.27 \\ \hline
\end{tabular}}
\caption{Ranking Results on Temporal Testsets}
\label{tab:temporal_result}
\end{table*}

\begin{table}[htbp!]
\centering
\resizebox{0.5\textwidth}{!}{
\begin{tabular}{|c|c|c|c|c|c|}
\hline
Metric & MRR & MP@1 & MP@3 & MP@5 & MP@10 \\ \hline
BW2V & 0.25 & 0.20 & 0.27 & 0.29 & 0.35 \\ \hline
TW2V & 0.00 & 0.00 & 0.00 & 0.00 & 0.00 \\ \hline
AW2V & 0.17 & 0.11 & 0.18 & 0.24 & 0.33 \\ \hline
DW2V & 0.12 & 0.11 & 0.11 & 0.13 & 0.14 \\ \hline
CW2V & \textbf{0.31} & \textbf{0.24} & \textbf{0.35} & \textbf{0.39} & \textbf{0.46} \\ \hline
\end{tabular}}
\caption{Ranking Results on Spatial Testset}
\label{tab:spatial_result}
\end{table}

\subsection{Quantitative Evaluation}
We also perform a quantitative evaluation of the condition-specific embeddings on the task of semantic equivalence across condition values. The joint embedding training is to bring the time- or location-specific embeddings to the same vector space so that they are comparable. Therefore, one key aspect of embeddings that we can evaluate is their semantic equivalence over time and locations. Two datasets with temporally- and spatially- equivalent word pairs were used for this part.

\subsubsection{Dataset}

\textbf{Temporal dataset}.  \citeauthor {yao2018dynamic} created two temporal testsets to examine the ability of the derived word embeddings to identify lexical equivalents over time \cite{yao2018dynamic}.  For example, the word \emph{Clinton-1998} is semantically equivalent to the word \emph{Obama-2012}, since Clinton was the US president in 1998 and Obama took office in 2012.

The first temporal testset was built on the basis of public knowledge about famous roles at different times such as the U.S. presidents in history. It consists of $11,028$ word pairs which are semantically equivalent across time.
For a given word in specific time, we find the closest neighbors of the time-dependent embedding in a target year. The neighbors are taken as its equivalents at the target time.

The second testset is about technologies and historical events. Annotators generated $445$ conceptually equivalent word-time pairs such as twitter-2012 and newspaper-1990. Here the equivalence is functional considering that  Twitter played the role of an information dissemination platform in 2012 just as the newspaper did in 1990.

\textbf{Spatial dataset}. To evaluate the quality of location-specific  embeddings, we created a dataset of $714$ semantically equivalent word pairs in different locations based on public knowledge. For example, the capitals of different countries have a semantic correspondence, resulting in the word \emph{Ottawa-Canada} that refers to the word \emph{Ottawa} for Canada to be equivalent to the word \emph{Dublin-Ireland} that refers to the word \emph{Dublin} used for Ireland. Two annotators {chose a set of categories such as capitals and governors} and independently came up with equivalent word pairs in different regions. Later they went through the word pairs together and decided the one to include. We will release this dataset upon acceptance.

\subsubsection{Evaluation metric}
In line with prior work \cite{yao2018dynamic}, we use  two evaluation metrics---mean reciprocal rank (MRR) and mean precision@k (MP@K)---to evaluate semantic equivalence on both temporal and spatial datasets.\\
\noindent \textbf{MRR}. For each query word, we rank all neighboring words in terms of their cosine similarity to the query word in a given condition, and identify the rank of the correct equivalent word. We define $r_{i}$ as the rank of the correct word of the $i$-th query, and MRR for $N$ queries is defined as $$\text{MRR}=\frac{1}{N}\sum\limits_{i=1}^{N}\frac{1}{r_{i}}.$$ Note that we only consider the top 10 words, and the inverse rank $1/r_i$ of the correct word is set as 0 if it does not appear among the top 10 neighbors. \\
\noindent \textbf{MP@K}. For each query, we consider the top-K words closest to the word in terms of cosine similarity in a given condition. If the correct word is included, we define the precision of the $i$-th query $\text{P@K}_{i}$ as 1, otherwise, $\text{P@K}_{i}=0$. MP@K for $N$ queries is defined as $$\text{MP@K}=\frac{1}{N}\sum\limits_{i=1}^{N}P@K_{i}.$$

\subsubsection{Results}

\noindent\textbf{Temporal testset}. We report the ranking results on the two temporal testsets in Table~\ref{tab:temporal_result}, and report results on the spatial testset in Table~\ref{tab:spatial_result}. Our condition-specific word embedding is denoted as CW2V in the tables.
{In  the temporal testset 1, our model is consistently better than the three baselines BW2V, TW2V and AW2V, and is comparable to DW2V in all metrics.}

{In the temporal tesetset 2, CW2V outperforms BW2V, TW2V and AW2V in all metrics and is comparable to DW2V with respect to precision in the top 1 and top 3 words, but falls behind DW2V in MP@5 and MP@10.} This lower performance may actually be a misrepresentation of its actual performance, since the word pairs in testset 2 are generated based on human knowledge and is potentially more subjective than testset 1. 

As an illustration, consider the case of \emph{website-2014} in testset 2. Our embeddings show \emph{abc}, \emph{nbc}, \emph{cbs} and \emph{magazine} as semantically similar words in 1990. These words are reasonable results since a website acts as a news platform just like TV broadcasting companies and magazines. The ground truth neighbor of \emph{website-2014} is the word \emph{address}.
Another example is \emph{bitcoin-2015}. The  semantic neighbors of our embeddings are \emph{currency}, \emph{monetary} and \emph{stocks} in 1992. These words are semantically similar to \emph{bitcoin} in the sense that \emph{bitcoin} is cryptocurrency and a form of electronic cash. However, the ground truth is \emph{investment} in the testset.

\noindent\textbf{Spatial testset}. Considering the evaluation on the spatial testset in Table~\ref{tab:spatial_result}, our condition-specific embedding achieves the best performance in finding semantic equivalents across regions. We note that the approaches which align independently trained embeddings such as TW2V and AW2V have poor performance. Due to the disparity in word distributions across regions in the ICE corpus, words with high frequency in one region may seldom be seen in another region. These infrequent words tend to have low-quality embeddings. It hurts the accurate alignment between locations and further degrades the performance of location-specific embeddings. 

DW2V, the jointly trained embedding, does not perform well on the spatial testset. It puts alignment constraints on word embeddings between two regions to prevent major changes of word embeddings across regions. This may lead to an interference between regional embeddings especially in cases where there is a frequency disparity of the same word in different regional corpora. In such cases, the embedding of the frequent word in one region will be affected by the weak embedding of the same word occurring infrequently in another region. Our model decomposes a word embedding into three components: a condition-independent component, a condition vector, and a deviation vector. The condition vector for each region takes care of the regional disparity, while the condition-independent vectors are not affected. Therefore, our model is more robust to such disparity in learning conditioned embeddings.


\section{Conclusion}
\label{sec:conclusion}
We propose a model to enrich word embeddings with temporal and spatial information. Our model explicitly encodes lexical semantic properties into the geometry of the embedding, and is empirically shown to well capture the language evolution and change with time and locations. We leave it to future work, to explore concrete downstream applications, where these time- and location-sensitive embeddings can be fruitfully used.

\section*{Acknowledgments}
We would like to thank the anonymous reviewers for their constructive comments and suggestions. We also thank Danny Polyakov and Yuchen Li for data annotations.

\bibliographystyle{acl_natbib}
\bibliography{emnlp2020}

\section{Supplemental Material}
\label{sec:supplemental}

\subsection{Previous Works on Temporal Embeddings}
\label{sec:app_related}

We first introduce some notations to facilitate our discussion of time-specific embeddings in the remaining part. Suppose that the size of word vocabulary $W$ is $|W|$ and word embeddings are of dimension $m$. The embedding matrix at time $t$ is denoted as $\mathbf{V}^{(t)}\in\mathbb{R}^{m\times|W|}$, and $\mathbf{v}_{w,t}\in\mathbb{R}^{m}$ is the vector of word $w$ at time $t$.

Existing approaches on time-specific embeddings can be divided into three categories: alignment of independently trained embedding, joint training of embeddings at different times and contextualized representations as time-sensitive sense embeddings. The first type of approaches include \cite{kulkarni2015statistically}, \cite{hamilton2016diachronic} and \cite{zhang2016past}. They pre-trained multiple sets of embeddings $\{\mathbf{V}^{(t)}\}_{t}$ for different times $t$ independently. Then one set of embedding is projected  to the space of another set so that two sets of embeddings are comparable.

\citeauthor{kulkarni2015statistically} assumes that word vector spaces at different times are equivalent under linear transformations, and learns an alignment matrix between two sets of embeddings \cite{kulkarni2015statistically}. Furthermore, it assumes the local structure preservation in embeddings across time, and use word neighbors to learn the transformation matrix $\mathbf{R}$ for a word $w$ from time $t_{1}$ to $t_{2}$. Suppose that $\mathbf{v}_{w,t}$ is the embedding of word $w$ at time $t$, and $k\text{NN}(\cdot)$ gives the nearest words in the vector space.

\begin{small}
\begin{align*}
\mathbf{R}_{w,t_{1},t_{2}} = \argmin\limits_{\mathbf{Q}}\sum\limits_{w_{i}\in k\text{NN}(\mathbf{v}_{w,t})}\lVert \mathbf{Q}\mathbf{v}_{w_{i},t_{1}}-\mathbf{v}_{w_{i},t_{2}}\rVert_{2}^{2}.
\end{align*}
\end{small}

Based on the same assumption of space equivalence under the linear transformation, \citeauthor{hamilton2016diachronic} finds an alignment matrix $\mathbf{R}^{(t)}\in\mathbb{R}^{m\times m}$ consisting of basis vectors so that the mean square error between the transformed embedding at time $t$ and embedding at time $t+1$ is minimized \cite{hamilton2016diachronic}.
\begin{align*}
\mathbf{R}^{(t)} = \argmin\limits_{\mathbf{Q}^{T}\mathbf{Q}=I}\lVert \mathbf{Q}\mathbf{V}^{(t)}-\mathbf{V}^{(t+1)} \rVert_{F}.
\end{align*}
\citeauthor{zhang2016past} finds the linear transformation using anchor words whose meaning remains stable across time \cite{zhang2016past}. It requires expert knowledge to find these stable words, which limits its application to general corpora.
\begin{figure*}[ht]
\centering
\begin{minipage}[c]{0.56\textwidth}
\centerline{\includegraphics[width=\linewidth]{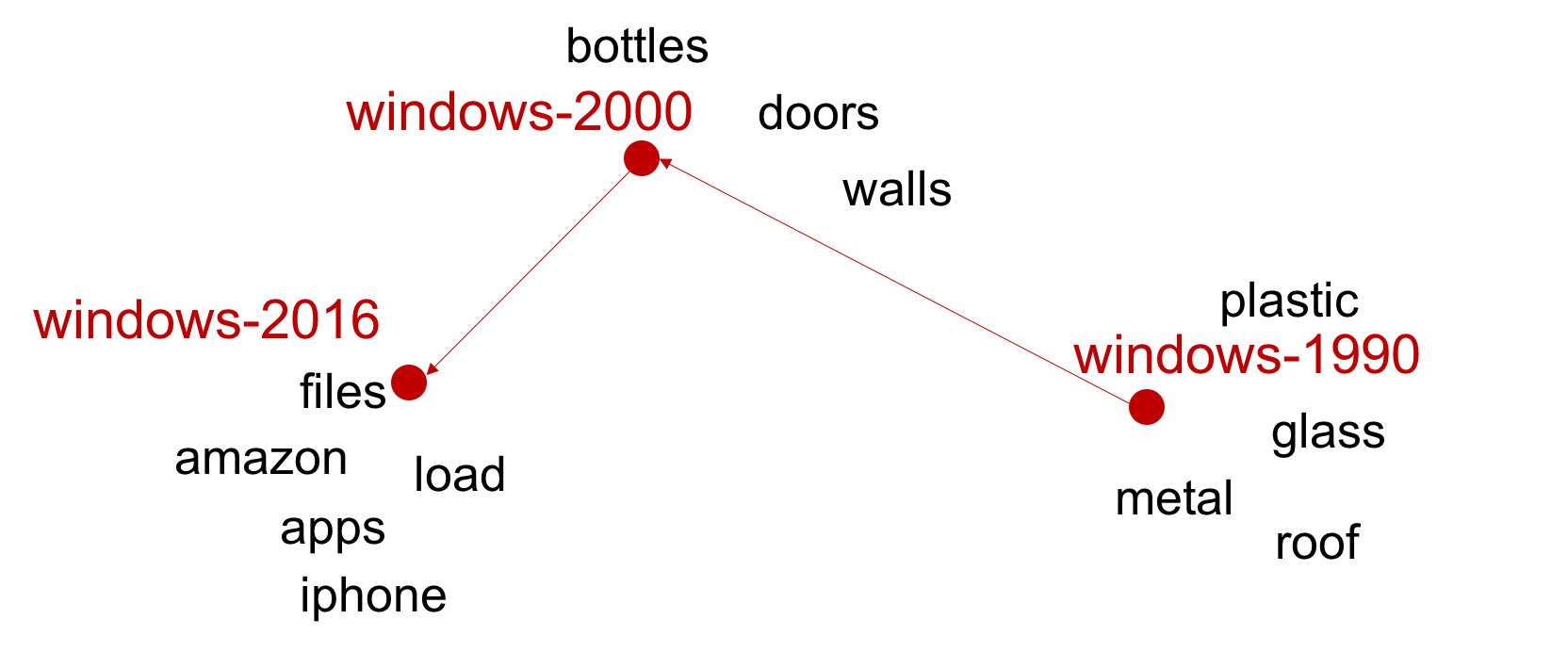}}
\centerline{\small{(a) Word ``windows''}}
\label{fig:windows_trajectory}
\end{minipage}
\begin{minipage}[c]{0.42\textwidth}
\centerline{\includegraphics[width=\linewidth]{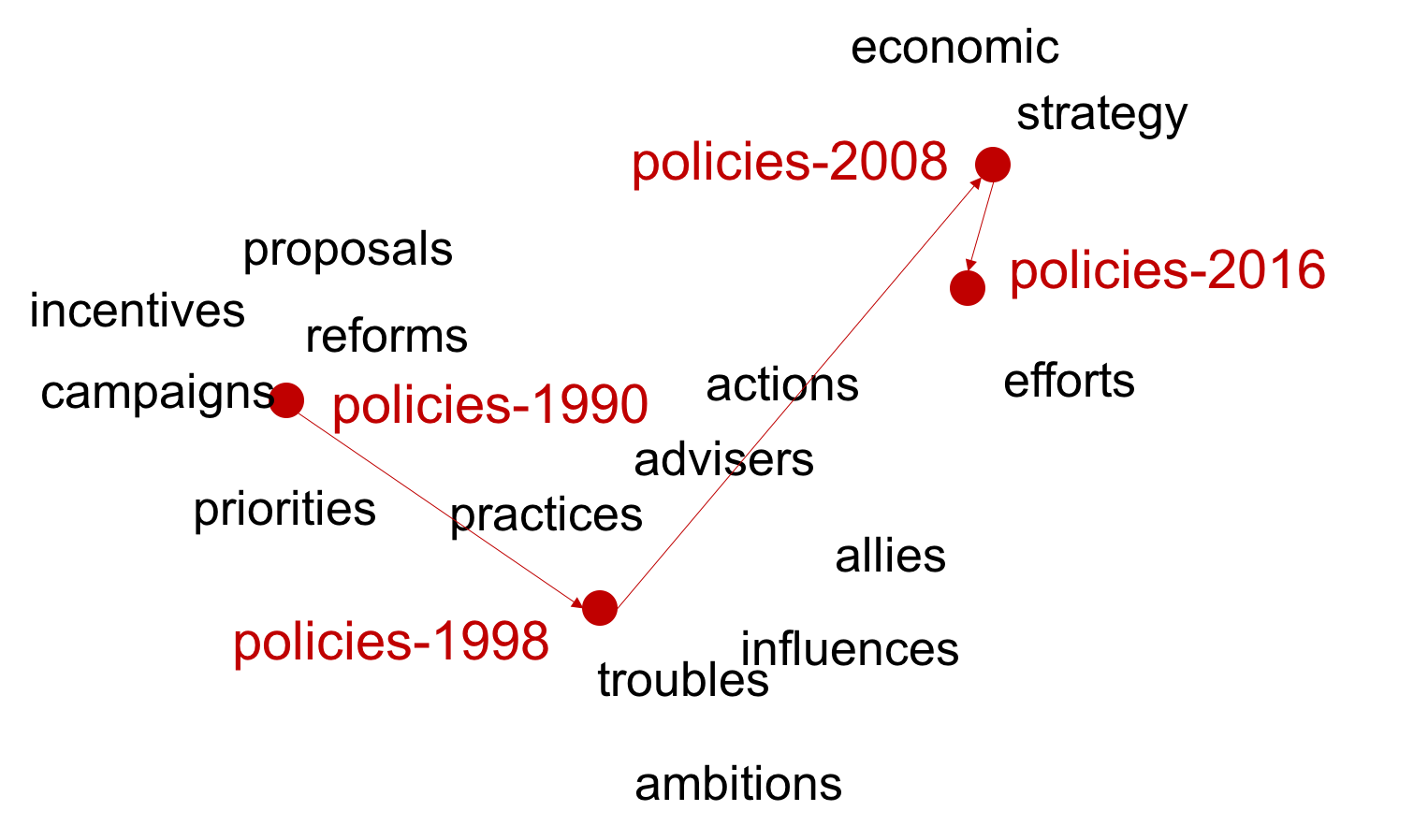}}
\centerline{\small{(b) Word ``policies''}}
\label{fig:policies_trajectory}
\end{minipage}
\caption{The trajectory of word embeddings over time.}
\label{fig:app_trajectory}
\end{figure*}

Different from aligning independently pre-trained embeddings, joint learning of time-stamped embeddings are shown to better capture semantic changes across time.
\cite{bamler2017dynamic,yao2018dynamic}. \citeauthor{bamler2017dynamic} uses a probabilistic language model to capture latent trajectories of word vectors across time \cite{bamler2017dynamic}. Their model is based on Bayesian skip-gram model, a probabilistic variant of word2vec. It learns embeddings $\mathbf{V}^{(t)}$ and context embeddings $\mathbf{U}^{(t)}$ at each time $t$. To align word embeddings across time, they add Gaussian assumption on the evolution of embeddings from time $t$ to $t+1$. They assume that the probability of $\mathbf{U}^{(t+1)}$ conditioned on $\mathbf{U}^{(t)}$, $p(\mathbf{U}^{(t+1)}|\mathbf{U}^{(t)})$,  follows Gaussian distribution. This Gaussian constraint prevents embeddings from growing large and enforces smooth vector trajectories.

\citeauthor{yao2018dynamic} used matrix factorization to learn an embedding matrix $\mathbf{V}^{(t)}$ and a context embedding matrix $\mathbf{U}^{(t)}$ from PPMI matrix $\mathbf{Y}^{(t)}$ with alignment constraints \cite{yao2018dynamic}.
\begin{align*}
&\mathbf{U}^{*}, \mathbf{V}^{*} = \argmin\limits_{\mathbf{U}^{(t)}, \mathbf{V}^{(t)}}\frac{1}{2}\sum\limits_{t=1}^{T}\lVert \mathbf{Y}^{(t)}-\mathbf{V}^{(t)}{\mathbf{U}^{(t)}}^{T}\rVert_{F}^{2} \\
&+\frac{\gamma}{2}\sum\limits_{t=1}^{T}\lVert \mathbf{V}^{(t)}-\mathbf{U}^{(t)}\rVert_{F}^{2}\\
&+ \frac{\lambda}{2}\sum\limits_{t=1}^{T}\lVert \mathbf{V}^{(t)}\rVert_{F}^{2} + \frac{\tau}{2}\sum\limits_{t=2}^{T}\lVert \mathbf{V}^{(t-1)}-\mathbf{V}^{(t)}\rVert_{F}^{2} \\
&+ \frac{\lambda}{2}\sum\limits_{t=1}^{T}\lVert \mathbf{U}^{(t)}\rVert_{F}^{2} + \frac{\tau}{2}\sum\limits_{t=2}^{T}\lVert \mathbf{U}^{(t-1)}-\mathbf{U}^{(t)}\rVert_{F}^{2},
\end{align*}
where terms $\lVert \mathbf{V}^{(t-1)}-\mathbf{V}^{(t)}\rVert_{F}^{2}$ and $\lVert \mathbf{U}^{(t-1)}-\mathbf{U}^{(t)}\rVert_{F}^{2}$ are alignment constraints on embeddings in neighboring time periods.

The third category of temporal embedding models are built upon pre-trained language models such as BERT \cite{devlin2018bert}. These models are pre-trained on large corpus to learn representations for words in a given context, which can be taken as the sense embedding. Time-specific word semantics are treated as different senses of words, and thus are represented by the contextualized representations \cite{hu2019diachronic,DBLP:conf/acl/GiulianelliTF20}.

\begin{figure}[ht]
\centering
\begin{minipage}{0.48\textwidth}
\centerline{\includegraphics[width=\linewidth]{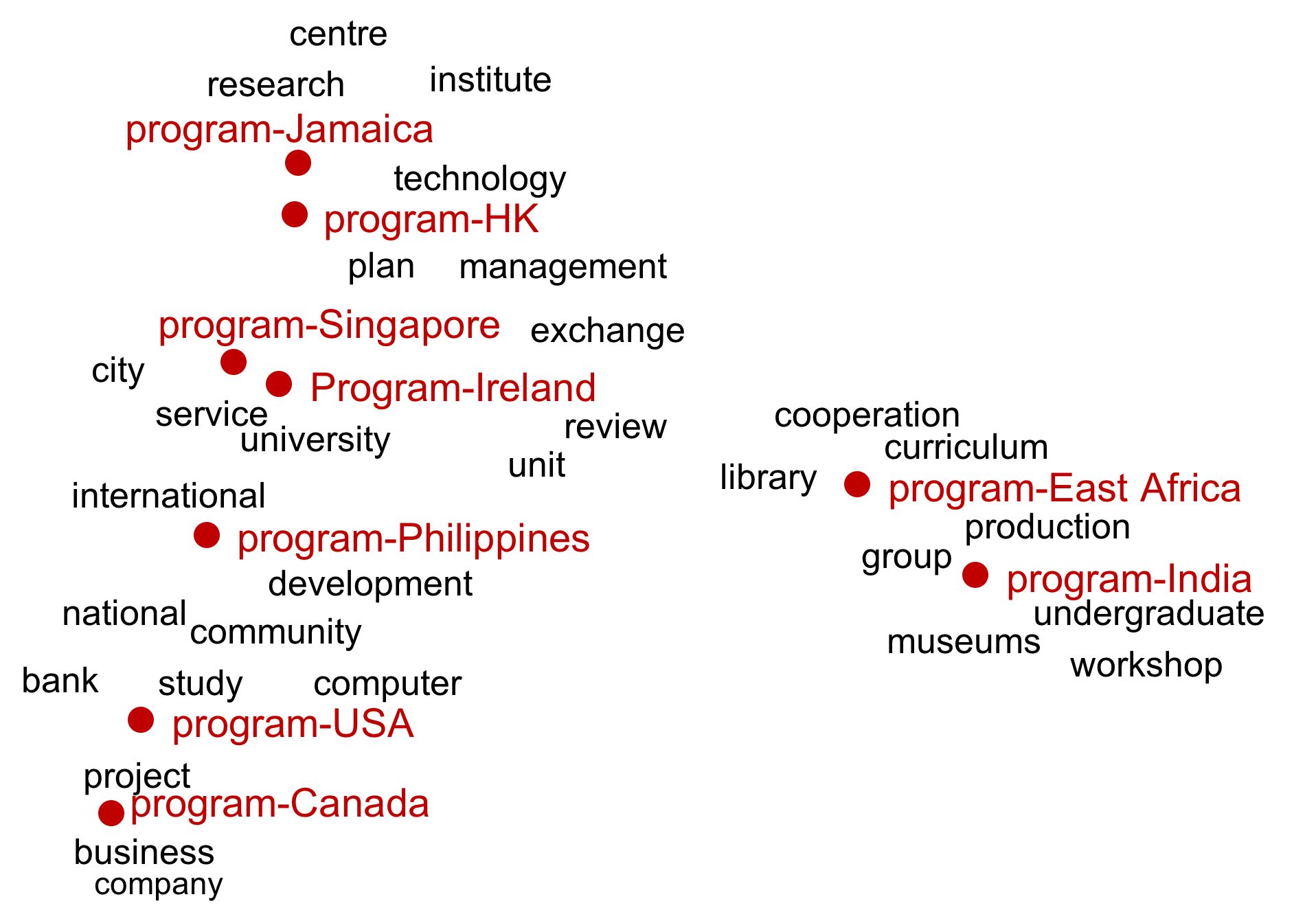}}
\end{minipage}
\caption{The trajectory of word embeddings over locations.}
\label{fig:app_region_trajectory}
\end{figure}
\subsection{Model and Optimization Problem}
\label{sec:app_derivation}

A model of static word embeddings is proposed by \citeauthor{arora2015rand}, and provides a unified understanding of a group of embedding models including pointwise mutual information (PMI) method, word2vec and GloVe \cite{arora2015rand}. It reveals that all these models train embeddings to estimate word co-occurrences in the training corpus.
Suppose that $\mathbb{P}_{s}(w_{1},w_{2})$ is the co-occurrence probability of words $w_{1}$ and $w_{2}$ in context window of size $s$, $\mathbf{v}_{w_{1}}$ and $\mathbf{v}_{w_{2}}$ are word vectors. Their model states that

\begin{small}
\begin{align}
\label{eq:arora}
\log \mathbb{P}_{s}(w_{1},w_{2}) = \frac{\lVert \mathbf{v}_{w_{1}}+\mathbf{v}_{w_{2}}\rVert_{2}^{2}}{2d} - 2\log Z + \gamma \pm \epsilon,
\end{align}
\end{small}

where $d$ is the embedding dimension, $Z$ is a constant, $\gamma=\log\left(\frac{s(s-1)}{2}\right)$ and $\epsilon$ is an error term. We consider the window size $s$ to be a  constant.
Since the coefficient $\frac{1}{d}$ can be absorbed as a constant scale of the word vectors, the model suggests the approximation of the logarithm of word co-occurrence probability below:
\begin{align*}
\log\mathbb{P}(w_{1},w_{2}) = \frac{1}{2}\lVert \mathbf{v}_{w_{1}}+\mathbf{v}_{w_{2}}\rVert_{2}^{2} + \tau,
\end{align*}
where constant $\tau= - 2\log Z + \gamma$.

We propose a model for condition-specific word embeddings in condition $c$, where $c$ can be time or location.
\begin{align*}
\log\mathbb{P}_{c}(w_{1}, w_{2})=\frac{1}{2}\lVert \mathbf{v}_{w_{1},c}+\mathbf{u}_{w_{2},c}\rVert_{2}^{2} + \tau.
\end{align*}

Since we assume that $\mathbf{v}_{w,c}=\mathbf{v}_{w}\odot \mathbf{q}_{c}+\mathbf{d}_{w,c}$, we can substitute $\mathbf{v}_{w_{1},c}$ with $\mathbf{v}_{w_{1}}\odot \mathbf{q}_{c}+\mathbf{d}_{w_{1},c}$, and substitute $\mathbf{u}_{w_{2},c}$ with $\mathbf{u}_{w_{2}}\odot \mathbf{q}_{c}+\mathbf{d}_{w_{2},c}'$. We then have
\begin{small}
\begin{align}
\nonumber
&\log\mathbb{P}_{c}(w_{1},w_{2}) \\
\nonumber
&= \frac{1}{2}\lVert (\mathbf{v}_{w_{1}}\odot \mathbf{q}_{c}+\mathbf{d}_{w_{1},c})+(\mathbf{u}_{w_{2}}\odot \mathbf{q}_{c}+\mathbf{d}_{w_{2},c}')\rVert^{2} + \tau, \\
\nonumber
&= \frac{1}{2}\left(\lVert \mathbf{v}_{w_{1}}\odot \mathbf{q}_{c}+\mathbf{d}_{w_{1},c}\rVert^{2} + \lVert \mathbf{u}_{w_{2}}\odot \mathbf{q}_{c}+\mathbf{d}_{w_{2},c}\rVert^{2}\right) \\
\nonumber
&+ (\mathbf{v}_{w_{1}}\odot \mathbf{q}_{c}+\mathbf{d}_{w_{1},c})^{T}(\mathbf{u}_{w_{2}}\odot \mathbf{q}_{c}+\mathbf{d}_{w_{2},c}') + \tau, \\
\nonumber
&= \frac{1}{2}(\lVert \mathbf{v}_{w_{1},c}\rVert^{2}+\lVert \mathbf{u}_{w_{2},c}\rVert^{2} + 2\tau) +\\ 
&(\mathbf{v}_{w_{1}}\odot \mathbf{q}_{c}+\mathbf{d}_{w_{1},c})^{T}(\mathbf{u}_{w_{2}}\odot \mathbf{q}_{c}+\mathbf{d}_{w_{2},c}').
\end{align}
\end{small}
Here $\frac{1}{2}\lVert \mathbf{v}_{w_{1},c}\rVert^{2}$ can be taken as the bias related to words $w_{1}$ under condition $c$. Similar to GloVe, we introduce a bias term $b_{w_{1},c}$, and $b_{w_{1},c}=\frac{1}{2}\lVert \mathbf{v}_{w_{1},c}\rVert^{2}+\tau$. Similarly, we define bias $b_{w_{2},c}'=\frac{1}{2}\lVert \mathbf{v}_{w_{2},c}\rVert^{2}+\tau$.

The logarithm of co-occurrence probability is:
\begin{align}
\nonumber
&\log\mathbb{P}_{c}(w_{1},w_{2}) \\
\nonumber
&= (\mathbf{v}_{w_{1}}\odot \mathbf{q}_{c}+\mathbf{d}_{w_{1},c})^{T}(\mathbf{v}_{w_{2}}'\odot \mathbf{q}_{c}+\mathbf{d}_{w_{2},c}') \\
&+ b_{w_{1},c} + b_{w_{2},c}'.
\end{align}

Our model has two sets of basic word vectors $\{\mathbf{v}_{w}\}$ and $\{\mathbf{u}_{w}\}$, two sets of deviation embeddings $\{\mathbf{d}_{w,c}\}$ and $\{\mathbf{d}_{w,c}'\}$, and two sets of bias terms $\{b_{w,c}\}$ and $\{b_{w.c}'\}$ for the vocabulary and context vocabulary respectively. All of these parameters are trained to minimize the difference between the real word co-occurrences and the estimated values.

\subsection{Word Embedding Trajectory}
\label{sec:app_trajectory}
The trajectories of word \emph{windows} and \emph{policies} across time are shown in Fig.~\ref{fig:app_trajectory} (a) and (b). As we can see, \emph{windows} was commonly used as an opening in houses to allow light and air before the 20th century since its embedding has neighbors such as \emph{glass} and \emph{walls}. Recently it also refers to an operating system developed by Microsoft given its neighbors \emph{files} and \emph{load}. As for the word \emph{policies}, it was relevant to \emph{campaigns} and \emph{reforms} in 1990, since many campaigns and reforms were launched in that cold-war era. Its meaning has shifted to be relevant to \emph{economic} over time.

In Fig.~\ref{fig:app_region_trajectory}, we plot the location-specific neighbors of word \emph{program}. We note that \emph{program} is a polysemous words with senses including \emph{project}, \emph{software} and \emph{curriculum}.  People in one region use it to refer to a sense that is different from another sense used in another region. The different senses can be inferred from its region-specific neighbors. For example, the \emph{program} is meant as projects or business in Canada, while it is also related to computer software in USA. East Africa and India use it to refer to curriculum in the education domain.




\end{document}